# Template-as-Ontology: Configurable Synthetic Data Infrastructure for Cross-Domain Manufacturing AI Validation

*Grama Chethan*

Siemens Digital Industries Software, Plano, TX 75024, United States

grama.chethan@siemens.com

ORCID: 0000-0000-0000-0000

***Abstract***— Large language model (LLM)-based AI agents deployed in manufacturing environments require populated, schema-correct data for validation, yet production MES data is proprietary, privacy-encumbered, and vendor-specific. This paper introduces the *Template-as-Ontology* principle: a single Python configuration module (700–770 lines, 45 validated exports) serves simultaneously as the specification for a time-stepped manufacturing simulator and as the runtime domain schema for AI analytics tools, producing alignment by construction rather than integration. We formally define the domain template as a typed relational configuration schema and prove that structural alignment between simulation and tool layers is guaranteed by single-source consumption. A five-layer pipeline—simulation, PostgreSQL, CDC/Iceberg lakehouse, star schema, and 12 parameterized AI tools—generates causally coherent, MES-shaped data spanning 66 entity types across four operational domains mapped to ISA-95/IEC 62264. We validate the architecture with six industry templates (aerospace, pharma, automotive, electronics, beverages, warehousing) running on identical framework code. Calibration experiments (60 runs, 10 seeds per template) confirm parametric controllability: observed KPIs fall within configured ranges across all templates. A controlled hallucination experiment (72 tool invocations, Qwen3-32B) demonstrates that ontology-constrained parameters eliminate tool-parameter fabrication (0% constrained vs. 43% unconstrained hallucination rate for the evaluated model, Fisher's exact test $p < 10^{-12}$); the 0% constrained rate is an architectural guarantee that holds for any model. The framework provides a reusable data layer for discrete manufacturing AI validation.

---

## I. INTRODUCTION

Validating AI systems for manufacturing requires data that is structured as a real Manufacturing Execution System (MES) would produce it: work orders with state machines, inspection results tied to measurement values, non-conformance reports (NCRs) linked to root causes, and material genealogy traceable to suppliers. Such data is locked inside proprietary MES systems, entangled with personally identifiable operator information, and structured in vendor-specific formats that no two plants share. It is not CDC-ready, not reproducible, and not available for the cross-domain testing that any portable AI platform requires.

A companion paper [1] identified the *semantic training gap*—the structural disconnect between how AI systems acquire domain vocabulary through training and how manufacturing operations define meaning through ontological relationships. That work demonstrated a three-operation interface contract (*resolve* → *contextualize* → *annotate*) that makes every AI tool call grounded and traceable, answering *why* ontology must be embedded in the tool layer. This paper answers *how*: how to build the data layer that makes such an architecture testable.

Ontology without data is a schema with nothing to query. An interface contract with no records behind it is a type system for an empty database. Before validating that an AI agent resolves station identifiers correctly, station data must exist. Before testing that failure codes are domain-appropriate, NCR records must exist. Before verifying cross-domain portability, data from multiple domains must be available simultaneously and reproducibly.

The solution is a time-stepped simulator that generates realistic, schema-correct, CDC-ready manufacturing data from a single domain configuration file. The critical architectural decision—*Template-as-Ontology*—is that the configuration file defining the simulated factory *is* the domain schema for the AI tools. One file, two roles, complete alignment by construction. The template is the portable artifact: a single Python module carrying all domain semantics, consumed by the simulator to generate data and by the AI tools to constrain queries. We use the term "ontology" in the Gruber [20] sense of "a specification of a conceptualization"—specifically, a typed relational configuration schema with domain constraints, rather than a formal OWL/RDF ontology with inference capability (see Definition 1, Section III-A).

This paper makes three contributions:

1. **Template-as-Ontology**: the principle that a single domain configuration module serves as both simulation specification and AI tool domain schema, producing alignment by construction (Property 1). We formally define the template structure, map it to ISA-95/IEC 62264 (Table II), and demonstrate this across six manufacturing domains with a 45-export interface validated at load time.
2. **A controlled hallucination experiment** showing that ontology-constrained tool parameters eliminate tool-parameter fabrication (0% vs. 43% hallucination rate across 72 queries and six templates, Fisher's exact test $p < 10^{-12}$, Cohen's $h = 1.46$), quantifying the value of the approach for AI tool validation.
3. **An end-to-end synthetic data pipeline**—from time-stepped simulation through CDC to a star schema and 12 AI analytics tools—producing causally coherent, MES-shaped data where the same configuration drives every layer.

## II. RELATED WORK

### A. Manufacturing Simulation

Discrete-event simulation (DES) of production systems is well-established, with commercial tools (Tecnomatix Plant Simulation, FlexSim, AnyLogic) providing high-fidelity process modeling [2]. Negri et al. [3] identify a gap between simulation models and operational data systems: simulation tools model material flow and resource contention but produce proprietary outputs (KPI logs, animation traces)—not CDC-ready transactional records structured as MES entities. The CMSD specification [4], developed by SISO/NIST, defines an interchange format for simulation inputs and outputs but does not generate operational records. Banks et al. [31] distinguish between time-stepped simulation (fixed-increment clock advance) and pure event-driven DES (next-event time advance). Our framework uses a time-stepped architecture (1-minute ticks), which trades sub-minute temporal precision for architectural simplicity and deterministic reproducibility (see Section IV-A). This is a deliberate design choice: the framework simulates *the data that processes produce*, not the processes themselves.

### B. Synthetic Data Generation

Statistical generators (Faker, Synthetic Data Vault [5], Gretel [6]) produce structurally correct records matching a target schema but cannot produce causal coherence: the relationship between a work order, its operations, the equipment that executed them, the inspection that followed, and the NCR that resulted. Process mining log generators (PLG2 [7], BPIC synthetic event logs) produce causally coherent event sequences from process models but generate event logs (activity, timestamp, case ID), not full MES entity graphs. Xu et al. [8] survey synthetic data methods for time-series applications but do not address multi-entity transactional data. Our system produces 66 entity types across four domains with full referential integrity and zero referential-integrity violations (see Section V-E).

### C. LLM Tool-Use and Grounding

Recent work on LLM function calling demonstrates that models hallucinate tool parameters at significant rates when schemas are unconstrained. Patil et al. [9] show that API-specific fine-tuning (Gorilla) reduces but does not eliminate hallucination for domain-specific tools. Schick et al. [10] demonstrate that tool-augmented language models can learn tool use, but evaluation uses general-purpose APIs, not MES queries where plausible identifiers map to nothing. Standard frameworks (OpenAI function calling, Anthropic tool use [11]) enforce syntactic constraints (types, enums); our work extends this to semantic constraints dynamically projected from a loaded domain schema. NeMo Guardrails [12] operates on model text output via post-hoc validation; our approach validates tool-call parameters *before* execution, providing the strongest possible constraint: structural impossibility of fabrication. Ji et al. [24] and Huang et al. [25] survey hallucination taxonomy broadly; we focus specifically on *tool-parameter fabrication*—the generation of plausible but non-existent identifiers in function-calling parameters (see Section V-C for precise definitions).

### D. Ontology and Domain Schemas for Industrial AI

Knowledge graphs and formal ontologies have been applied to manufacturing contexts, including ISA-95 formalization [13], [14], equipment maintenance reasoning [15], and quality analysis. Gruber [20] defines an ontology as "a specification of a conceptualization"; Guarino [32] further distinguishes lightweight ontologies (taxonomies, vocabularies) from formal ontologies (with axioms, inference, and consistency guarantees in OWL/SHACL). Pan et al. [16] survey the intersection of knowledge graphs and LLMs, identifying parameter grounding as open. Lemaignan et al. [17] developed MASON for manufacturing semantics. Biffl et al. [18] address semantic interoperability in cyber-physical production systems. Our Template-as-Ontology approach occupies a specific position on this spectrum: it is a *typed relational configuration schema* (Definition 1)—more structured than a flat vocabulary but less formal than an OWL ontology with axioms and inference. The key insight is that for runtime AI tool constraint validation, lightweight configuration with structural guarantees (Property 1) is more appropriate than formal inference, because the validation question is membership ("is S1 a valid station?") rather than subsumption ("is CNC Machining a subclass of Material Removal?").

## III. TEMPLATE-AS-ONTOLOGY ARCHITECTURE

### A. Formal Definitions

We begin by formally defining the template structure and the alignment property it guarantees.

***Definition 1 (Domain Template).*** A domain template $T$ is a tuple $T = (E, P, F, R, C)$ where:

- $E = \{e_1, ..., e_n\}$ is the equipment hierarchy, a forest of typed nodes at levels {Site, Area, WorkCenter, Unit, Instrument} with parent references obeying the ISA-95 containment constraint: Unit ⊂ WorkCenter ⊂ Area ⊂ Site;
- $P = \{p_1, ..., p_m\}$ is the product set, each with station routing $p_i.stations \subseteq S$, where $S = \{s \in$ keys(STATIONS)$\}$ is the station identifier set;
- $F = \{f_1, ..., f_k\}$ is the failure code catalog, with each $f_j.station \in S$ (failure codes are station-scoped);
- $R$ is the set of relational constraints: the station-to-work-center mapping $\sigma: S \to WC$ is injective (each station maps to exactly one work center); the failure-code-to-station mapping $\varphi: F \to S$ is surjective over the stations with quality gates; and for every product routing, each station in the routing must map to a work center with at least one available equipment unit: $\forall p_i \in P, \forall s \in p_i.\text{stations}: \sigma(s) \in WC \wedge |\{u \in E : u.\text{level} = \text{Unit} \wedge u.\text{parent} = \sigma(s)\}| \geq 1$ (referential integrity of product routings);
- $C = \{c_1, ..., c_{45}\}$ is the set of 45 named export constants, validated at load time (Listing 1).

This definition positions the template as a *typed relational configuration schema*—more structured than a flat key-value configuration (it enforces typed cross-references and containment hierarchies) but less formal than an OWL ontology (it lacks inference rules, subsumption hierarchies, and axiom-based consistency checking). We retain the term "ontology" in the Gruber [20] sense: the template is a specification of the conceptualization of a manufacturing domain, sufficient for runtime constraint validation.

***Property 1 (Alignment by Construction).*** For any template $T$ loaded by `load_template(T)`, the set of valid identifiers $V_{\text{sim}}$ consumed by the simulator equals the set $V_{\text{tool}}$ used by the AI tool parameter constraints:

$$V_{sim} = V_{tool} = S \cup P.part_numbers \cup F.nids \cup E.nids \cup ...$$

*Structural argument.* Both the simulator (via `config.STATIONS`, `config.PRODUCTS`, etc.) and the AI tool constraint generator (via the same `config.*` references) read from the same Python module object. The `load_template` function (Listing 1) sets these as module-level globals via `setattr`. No copy is made; both consumers hold references to the same dictionary objects. Therefore $V_{\text{sim}}$ and $V_{\text{tool}}$ are not merely equal by value but identical by reference. Adding a station to the template makes it simultaneously available to the simulator for data generation and to the tool schema for constraint validation. Removing a station removes it from both. Drift between the two sets is structurally impossible because there is only one set.

*Language-agnostic restatement.* The property depends on the reference-semantic design of the current Python implementation: both consumers share the same in-memory objects rather than independent copies. In a language with value semantics (e.g., Go structs, Rust owned types), the same guarantee would require an equivalent architectural mechanism—such as a shared configuration service, an immutable snapshot pattern, or a single configuration store from which both consumers read—to ensure that $V_{\text{sim}}$ and $V_{\text{tool}}$ remain identical. The essential invariant is single-source consumption, not reference identity per se.

### B. Data Model Overview and ISA-95 Mapping

The simulator produces operational synthetic data structured as a real MES would produce it. The data model spans 66 entity types across four domains: (1) *MES/MOM* (28 types)—work orders, operations, steps, material tracking units, operator assignments, equipment usage, with state machines governing lifecycle transitions; (2) *BOP/ERP* (14 types)—process plans, planned operations, BOMs, tool requirements, material consumption specifications; (3) *Quality* (12 types)—inspection operations, characteristic specifications (variable and attributive), inspection samples and values, defect types, NCRs, corrective actions (CAPA); (4) *Master Data* (12 types)—ISA-95 equipment hierarchy (Site → Area → WorkCenter → Unit), certifications, skills, tool definitions, suppliers, shift schedules, SPC control limits. Every entity supports dual serialization: snake_case for PostgreSQL/CDC and PascalCase for OData/API consumption. The dual serialization is aspirational for MES interoperability rather than demonstrated against a production sys-tem; its primary current role is ensuring the generated data structurally matches real MES APIs.

Table II maps the framework's entities to ISA-95 activity models and IEC 62264 object model elements.

**TABLE II**
**ISA-95/IEC 62264 ENTITY MAPPING**

| Framework Entity | ISA-95 Level | IEC 62264 Object Model | Activity Model | Coverage |
|---|---|---|---|---|
| Equipment (Site) | Level 4 (Enterprise) | Equipment / PhysicalAsset | — | Full |
| Equipment (Area) | Level 3 (Site/Area) | Equipment / WorkCenter parent | — | Full |
| Equipment (WorkCenter) | Level 3 (WorkCenter) | EquipmentClass: WorkUnit | Production | Full |
| Equipment (Unit) | Level 2 (Unit) | Equipment: individual asset | Production | Full |
| WorkOrder | Level 3 | ProductionSchedule / ProductionRequest | Production | Full |
| WorkOrderOperation | Level 3 | SegmentRequirement / SegmentResponse | Production | Full |
| Process (plan) | Level 3 | OperationsDefinition | Production | Full |
| Operation (planned) | Level 3 | OperationsSegment | Production | Full |
| MaterialTrackingUnit | Level 3 | MaterialLot / MaterialSublot | Inventory | Partial |
| BillOfMaterials | Level 3–4 | MaterialBill (BOM) | Production | Full |
| NonConformance | Level 3 | QualityTestResult (extension) | Quality | Full |
| InspectionOperation | Level 3 | TestSpecification | Quality | Full |
| QualityAction (CAPA) | Level 3 | OperationsEvent (corrective) | Quality | Full |
| EquipmentEvent | Level 2–3 | OperationsEvent (maintenance) | Maintenance | Partial |
| Shift / WorkCalendar | Level 3 | PersonnelModel / WorkSchedule | Production | Partial |
| ISA-88 Procedural Model | Level 1–2 | BatchControl / Recipe | Production | Out of scope |
| Lot/Batch Tracking | Level 3 | MaterialLot split/merge | Inventory | Out of scope |
| Maintenance Schedule | Level 3 | MaintenanceRequest | Maintenance | Out of scope |

The framework covers ISA-95 Production and Quality activity models at Level 3 with full entity support. Inventory (partial: serial-level MTU tracking but no lot split/merge) and Maintenance (partial: events only, no scheduled work orders) have limited coverage. ISA-88 procedural models and advanced maintenance scheduling are explicitly out of scope.

### *C. Template Interface Specification*

The core architectural claim is that a single Python module —700 to 770 lines—defines everything about a manufacturing domain: equipment hierarchy, products, station routings, process plans, inspection plans, failure codes, operator certifications, tool definitions, and regulatory context. This is not a flat key-value configuration but a typed relational configuration (Definition 1) where stations reference work centers, products reference station sequences, failure codes are scoped to stations, and inspection plans are linked to operations.

The interface is enforced: every template must export exactly 45 named constants, validated at load time (Listing 1). The 10 functional categories provide complete coverage: plant configuration, equipment hierarchy, products, materials, quality, process parameters, workforce, tooling, step templates, and change management.

***Listing 1.*** *Required template exports (45 constants, validated at load time).*

```
REQUIRED_EXPORTS = [
    "PLANT_CODE", "PLANT_NAME", "SHIFTS", "OPERATING_DAYS",
    "BREAK_DURATION_MIN", "WEEKLY_PM_HOURS",
    "TARGET_OEE_RANGE", "FIRST_PASS_YIELD_RANGE", "AVG_WIP_RANGE",
    "OPERATORS_PER_SHIFT",
    "EQUIPMENT", "WORK_CENTER_UNITS", "PRODUCTS", "WORKING_DAYS_PER_YEAR",
    "STATIONS", "STATION_TO_WC",
    "RAW_MATERIALS", "FINISHED_MATERIALS", "PRODUCT_RAW_MATERIAL",
    "OPERATION_MATERIAL_CONSUMPTION",
    "SUPPLIERS", "FAILURE_CODES", "STATION_FAILURE_CODES",
    "PROCESS_PLANS", "INSPECTION_PLANS", "STATION_INSPECTION_PLANS",
    "NCR_DISPOSITIONS", "NCR_STATUS_DURATIONS", "CAPA_TRIGGER_RATE",
    "EQUIPMENT_DOWNTIME_PROB", "EQUIPMENT_DOWNTIME_DURATION_MIN",
    "ORDER_EXPEDITE_RATE", "BOP_REVISION_INTERVAL_DAYS",
    "CYCLE_TIME_VARIANCE", "DEFAULT_RANDOM_SEED",
    "CERTIFICATIONS", "STATION_CERTIFICATIONS",
    "SKILLS", "STATION_SKILLS",
    "TOOL_DEFINITIONS", "STATION_TOOLS",
    "STEP_TEMPLATES", "CHANGE_PACKAGE_RATE", "CHANGE_PACKAGE_PARAMS",
    "BOM_STATION_MATERIALS",
]

def load_template(template_id):
    """Load template and set all exports as module-level globals (setattr)."""
    mod = get_template_module(template_id)
    missing = [name for name in REQUIRED_EXPORTS if not hasattr(mod, name)]
    if missing:
        raise ValueError(f"Template {template_id!r} missing exports: {missing}")
    for name in REQUIRED_EXPORTS:
        setattr(config_module, name, getattr(mod, name))
    # Both simulator and AI tools read config.X — same object, zero drift
```

The template duality is best understood by comparing two domains. Listing 2 shows how the `STATIONS` dictionary differs between aerospace and pharma while preserving identical structure.

***Listing 2.*** *Station definitions: aerospace (left) vs. pharma (right). Same structure, different domain semantics.*

```
# aerospace.py                           # pharma.py
STATIONS = {                             STATIONS = {
    "S1": {                                  "S1": {
        "name": "CNC Machining",                 "name": "Dispensing",
        "work_center": "WC-CNC",                 "work_center": "WC-DISPENSE",
        "cycle_time_range_min": (120, 480),       "cycle_time_range_min": (20, 45),
        "setup_time_min": (30, 60),               "setup_time_min": (15, 30),
        "first_pass_yield": 0.95,                 "first_pass_yield": 0.99,
        "is_quality_gate": True,                  "is_quality_gate": True,
    },                                       },
    # S2-S6: Drilling, Riveting,             # S2-S6: Granulation, Blending,
    #   Bonding, NDT, Final Assembly          #   Compression, Film Coating, Packaging
}
```

### *D. Template Loading and Consumer Alignment*

The configuration loader reads the template module, validates its 45 exports, and replaces every module-level global so that all downstream consumers—simulator, seed generator, analytics tools, star schema builder—immediately see the new domain. The swap is a single function call: `load_template(template_id)`. No code changes, no redeployment, no schema migration.

This is the Template-as-Ontology duality in practice. When an analytics tool constrains its `station_id` parameter to `[S1, S2, ..., S6]`, those values come from `config.STATIONS`—the same dictionary the simulator used to generate the data. The `resolve` operation from the interface contract [1] resolves against the same source. One file, two consumers, zero drift. This structural guarantee (Property 1) is stronger than integration testing or schema synchronization: drift is impossible because there is only one source.

## IV. SIMULATION ENGINE AND DATA PIPELINE

### A. Time-Stepped Simulation Engine

The simulation engine is a *time-stepped* simulation with 1-minute tick resolution, not a pure discrete-event simulation (DES) in the Banks et al. [31] sense. In pure DES, the clock advances to the next scheduled event; in time-stepped simulation, the clock advances by a fixed increment and all pending events are evaluated at each tick. We chose time-stepped simulation for three reasons: (1) deterministic reproducibility—given the same seed, template, and duration, the engine produces byte-identical output; (2) architectural simplicity—no priority queue management, no event cancellation, no simultaneous-event tie-breaking; (3) calendar fidelity—the factory calendar (shifts, breaks, operating days) maps naturally to fixed-increment ticks.

The tradeoff is temporal precision. Events that would occur at sub-minute resolution (e.g., a 45-second inspection step) are rounded to the nearest minute. For the target use case—generating MES-shaped data for AI tool validation, not process optimization—this precision is sufficient. Cycle times in the six templates range from 20 minutes (pharma dispensing) to 720 minutes (aerospace bonding), making the 1-minute tick a ≤5% discretization error for even the shortest operations.

**Algorithm 1: Time-Stepped Simulation Main Loop**

**Input:** Template $T$, start time $t_0$, duration $D$ days, random seed $r$

**Output:** CDC-ready event stream $E$ to PostgreSQL

1: **procedure** RUN($T$, $t_0$, $D$, $r$)

2: rng ← Random($r$); $t \leftarrow t_0$; $t_{end} \leftarrow t_0 + D$

3: load_template($T$) *// Sets config.STATIONS, config.PRODUCTS, ... (Property 1)*

4: seed_data ← GenerateSeeds($T$, $t_0$) *// 30+ reference entity types*

5: WriteBatch(seed_data)

6: **while** $t < t_{end}$ **do**

7: **if** not IsWorkingTime($t$, $T$.SHIFTS, $T$.OPERATING_DAYS) **then**

8: $t \leftarrow t + 1$ min; **continue**

9: events ← []

10: *// Phase 1: Disruption handling*

11: events.extend(DrainPendingDisruptions($t$))

12: events.extend(ExpireActiveDisruptions($t$))

13: MaybeAutoInject($t$, rng) *// Exponential inter-arrival*

14: *// Phase 2: Daily order creation (at 06:00)*

15: **if** $t$.hour = 6 ∧ $t$.minute = 0 ∧ NewDay($t$) **then**

16: events.extend(GenerateDailyOrders($t$, $T$.PRODUCTS))

17: *// Phase 3: Four-gate operation scheduling*

18: **for each** queued operation $op$ **do**

19: **if** EquipAvail($op$) ∧ NoSupplyDelay($op$) ∧ UpstreamDone($op$) ∧ OperatorCertified($op$, $t$) **then**

20: StartOperation($op$, $t$); *// cycle_time ~ Uniform(T.STATIONS[s].ct_range)*

21: *// Phase 4: Complete operations → quality inspection*

22: **for each** active operation $op$ where $t \geq op$.end_time **do**

23: CompleteOperation($op$, $t$)

24: **if** Bernoulli(1 − $T$.STATIONS[$op$.station].FPY) **then**

25: ncr ← CreateNCR($op$, SampleFailureCode($op$.station, $T$.STATION_FAILURE_CODES))

26: events.append(ncr) *// NCR enters own lifecycle (Fig. 2)*

27: *// Phase 5: Periodic events (equipment, expedite, WIP snapshots)*

28: **if** $t$.minute mod 15 = 0 **then** events.extend(EquipmentEvents($t$))

29: **if** $t$.minute mod 30 = 0 **then** events.extend(NCRLifecycle($t$)); CAPALifecycle($t$)

30: WriteBatch(events)

31: $t \leftarrow t + 1$ min

32: **end while**

### *B. State Machines*

Work orders and operations follow defined state machines. A WorkOrder transitions through: Edit → New → Active → Complete (or Aborted). Transitions are triggered by simulation events: New → Active when the first operation starts; Active → Complete when the last operation completes. An Operation transitions through: New → Active → Complete (or Aborted). The Active → Complete transition triggers quality evaluation (FPY gate), which may spawn an NCR with its own lifecycle: New → InProcess → Pending Disposition → Closed. Each lifecycle has configured cadence: NCR status durations are template-defined (`NCR_STATUS_DURATIONS`), enabling domain-specific progression rates (pharma NCRs close faster than aerospace NCRs due to regulatory cadence differences).

### C. Stochastic Models

The simulator uses the following probability distributions, all parameterized by the loaded template:

- *Cycle times:* Uniform($ct_{\mathrm{lo}}$, $ct_{\mathrm{hi}}$) where ($ct_{\mathrm{lo}}$, $ct_{\mathrm{hi}}$) = `STATIONS[s].cycle_time_range_min`.
- *Setup times:* Uniform($su_{\mathrm{lo}}$, $su_{\mathrm{hi}}$) where ($su_{\mathrm{lo}}$, $su_{\mathrm{hi}}$) = `STATIONS[s].setup_time_min`.
- *Quality gate:* Bernoulli(1 − FPY) per station, independent draws.
- *Disruption inter-arrival:* Exponential(1/MTBF), memoryless, independent per disruption type, with a 5-minute floor.
- *Order volumes:* Deterministic via Bresenham-style fractional accumulation (not stochastic).

### D. Core Algorithms

*Fractional order accumulation.* Converting annual volume to daily work orders via rounding loses or gains units over time. A product at 850 units/year and 250 working days produces 3.4/day—rounding to 3 loses 100 units/year. The simulator uses Bresenham-style fractional accumulation: the remainder carries forward, producing sequences of 3, 4, 3, 4 that are exact over any window.

*Four-gate operation scheduling.* Each tick, every queued operation is evaluated against four conditions: (1) equipment is not busy or down, (2) no supply delay blocks the station, (3) upstream operation has completed, and (4) a certified operator is available in the current shift. When all four gates pass, the operation starts. Completion triggers a causal chain: the station's FPY is evaluated; failure generates an NCR with station-scoped failure codes, which may trigger a CAPA.

*Disruption composition.* Each of five disruption types maintains an independent exponential timer (`expovariate(1/mtbf)`). Key properties: memoryless inter-arrival, independent timers enabling concurrent disruptions at different stations, bounded concurrency, and a 5-minute floor preventing unrealistic rapid-fire draws.

*Template-adaptive query generation.* The most novel algorithm is in the analytics layer. When the star schema is built, SQL statements are generated dynamically from the loaded template via the `_dynamic_sql` function, which reads `config.STATIONS`, `config.PRODUCTS`, `config.FAILURE_CODES`, and `config.EQUIPMENT` to produce CASE statements mapping operational identifiers to analytical dimensions (Listing 3).

***Listing 3.*** *Template-adaptive SQL generation (`_dynamic_sql`, simplified). The function generates 10 CASE-statement blocks from the loaded template.*

```
def _dynamic_sql():
    wc_to_station = {s["work_center"]: sid for sid, s in config.STATIONS.items()}
    # Station dimension: one row per station from config
    rows = []
    for i, (sid, s) in enumerate(config.STATIONS.items(), 1):
        name = s["name"].replace("'", "''")
        fpy = s.get("first_pass_yield", 0.95)
        ct_lo, ct_hi = s.get("cycle_time_range_min", (60, 120))
        rows.append(f"('{sid}', '{name}', '{s['work_center']}', {i}, {fpy}, {(ct_lo+ct_hi)//2})")
    dim_station_sql = "INSERT INTO DimStation VALUES\n  " + ",\n  ".join(rows) + ";"
    # Operation-to-station CASE: maps PlannedWorkCenter to station_nid
    op_case = "\n  ".join(f"WHEN PlannedWorkCenter = '{wc}' THEN '{sid}'" for wc, sid in wc_to_station.items())
    # Defect-to-station CASE: maps failure code NId to station
    defect_case = "\n  ".join(f"WHEN DefectNId = '{fc['nid']}' THEN '{fc['station']}'" for fc in config.FAILURE_CODES)
    return {"dim_station": dim_station_sql, "op": op_case, "defect": defect_case, ...}
```

When the template switches from aerospace to pharma, the DimStation INSERT changes from `('S1', 'CNC Machining', ...)` to `('S1', 'Dispensing', ...)`, the operation CASE changes work center mappings, and the defect CASE changes from aerospace failure codes (bond-line void, burr, tool wear) to pharma failure codes (content uniformity, tablet hardness, coating defect). No SQL template is edited. No tool code changes.

### E. Five-Layer Data Pipeline

The template flows through every layer of the pipeline, from event generation to AI tool execution. Fig. 1 illustrates this architecture.

*Layer 1: Simulation.* The simulator reads station, product, process plan, and inspection plan configurations to generate work orders, advance operations, evaluate quality gates, and produce NCRs. The seed generator creates 30+ reference entity types from the same configuration.

*Layer 2: Operational database.* 40+ PostgreSQL tables receive events in real time via bulk-insert. Tables carry both `CreatedOn` and `ModifiedOn` timestamps for CDC readiness.

*Layer 3: CDC and Iceberg.* Mutable tables (6 tables: WorkOrder, WorkOrderOperation, NonConformance, ChangePackage, QualityAction, Equipment) use database triggers capturing INSERT/UPDATE/DELETE as JSONB payloads with OldRow/NewRow into a `_cdc_changelog` table. Append-only tables (5 tables: InspectionValue, InspectionSample, ActualConsumedMaterial, EquipmentEvent, Defect) use watermark polling with `CreatedOn ≥ high_water` and primary-key deduplication. The high watermark is held in memory for the duration of the sync process; it is not persisted to disk. If the sync process crashes mid-cycle, it restarts from the last successfully committed Iceberg snapshot timestamp, re-scanning the changelog or watermark window; primary-key deduplication ensures no duplicate rows are written. Both sync to Apache Iceberg tables via HTTP POST at 30-second intervals, writing Parquet files. The remaining 29+ tables are seed/reference data that do not change during simulation and are not CDC-tracked. Database triggers are used rather than WAL-based CDC for reasons discussed in Section IV-F.

*Layer 4: Star schema.* 23 analytics tables (14 dimensions, 8 facts, 1 bridge) are built via full rebuild triggered on demand. At current scale (15,000–54,000 rows), rebuild completes in under 2 seconds. This is where the template re-enters the pipeline: dynamic SQL generates CASE statements mapping operational identifiers to analytical dimensions. The full-rebuild approach is justified in Section IV-G.

*Layer 5: AI tools.* 12 parameterized SQL queries span five analytics domains (Table I). Each tool's parameters are constrained by the loaded template. Tools are exposed via the OpenAI-compatible function-calling protocol. The orchestrator validates inputs against the loaded template before executing, producing a clear error with the valid set rather than a silent empty result.

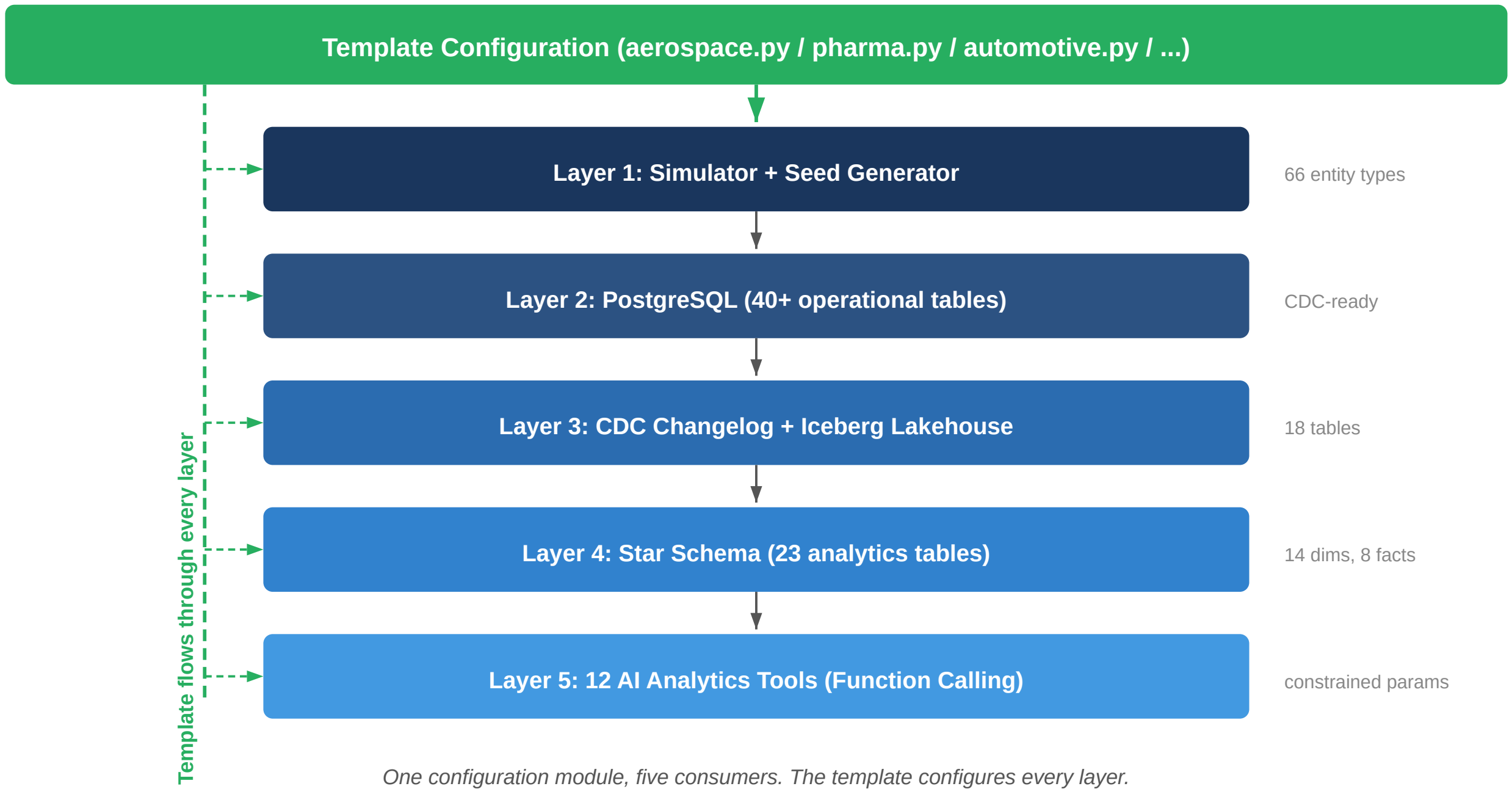

**Fig. 1.** Five-layer data pipeline architecture. The template configuration (green) flows through every layer, ensuring that the same domain semantics drive data generation, schema construction, and AI tool parameter constraints. Notation follows UML component diagram conventions: rectangles represent processing components, the dashed line represents the configuration dependency.

TABLE I
TWELVE AI ANALYTICS TOOLS SPANNING FIVE DOMAINS

| Domain | Tool | Key Parameters | Returns |
|---|---|---|---|
| Production | cycle_time_analysis | station_nid, program_code, time_range_days | Actual vs. planned variance, stddev, min/max per station |
| | first_pass_yield | station_nid, time_range_days, group_by | FPY trending with produced/scrapped/reworked counts |
| | oee_decomposition | station_nid, time_range_days | Availability × Performance × Quality per station |
| Quality | ncr_root_cause_pareto | station_nid, severity, time_range_days, top_n | Defect types ranked by count with cumulative Pareto % |
| | spc_violation_detection | characteristic_nid, station_nid | Out-of-spec counts, Cpk, mean/stddev per characteristic |
| | quality_action_status | status_filter, capa_type | Open/overdue CAPAs, sub-action counts |
| Materials | material_genealogy | serial_number, order_nid | Full trace: supplier → material → operation → product |
| | supplier_performance | supplier_code, time_range_days | Defect rate, lots received, COPQ per supplier |
| Eng. Change | change_impact_analysis | change_package_nid, time_range_days | Affected orders, related NCRs, quality impact |
| | engineering_change_velocity | time_range_days, change_type | Open-to-close duration, bottleneck identification |
| Operations | equipment_downtime_analysis | station_nid, time_range_days | Downtime causes, MTBF/MTTR by equipment |
| | production_status_summary | program_code, station_nid | WIP, on-hold, overdue, throughput per station |

*F. CDC Approach: Triggers vs. WAL-Based Alternatives*

The framework uses database triggers for CDC rather than WAL-based approaches. Table VIII compares three CDC mechanisms.

TABLE VIII
CDC MECHANISM COMPARISON

| Property | Trigger-Based (This Work) | WAL-Based (pgoutput) | Debezium (Kafka Connect) |
|---|---|---|---|
| Write amplification | High (JSONB row per change) | None (reads existing WAL) | None (reads existing WAL) |
| Schema coupling | Tight (trigger per table) | Loose (slot-level) | Loose (connector config) |
| Cross-table ordering | Not guaranteed | LSN-ordered | LSN-ordered |
| DDL capture | No | No (pgoutput) | Yes (with schema history) |
| Operational complexity | None (no external infra) | Low (replication slot) | High (Kafka + Connect + ZK) |
| Write latency overhead | ~0.1–0.3 ms/row | None | None |
| Suitable at scale | No (100K+ rows/day) | Yes | Yes |
| Infrastructure required | PostgreSQL only | PostgreSQL only | Kafka + Schema Registry |

Trigger-based CDC is appropriate for this framework because the simulator controls write volume (peak ~500 rows/tick, ~15,000 total for 30 days) and triggers require no external infrastructure. For production deployments exceeding 100K rows/day, WAL-based CDC via pgoutput or Debezium would be necessary. The ~0.1–0.3 ms per-row trigger overhead is negligible at current scale but would compound at higher volumes.

Trigger-based CDC preserves transactional consistency (the changelog write is in the same transaction as the original INSERT/UPDATE) but adds write amplification and does not provide cross-table ordering guarantees. For the controlled-write-volume synthetic data use case, these trade-offs are acceptable. We do not claim that trigger-based CDC is suitable for production MES environments; rather, we claim that the generated data is *structured for CDC* (with CreatedOn/ModifiedOn timestamps, primary keys, and change-tracking columns), making it compatible with any CDC mechanism.

*G. Star Schema Design and Full-Rebuild Justification*

The star schema comprises 23 analytics tables: 14 dimensions (DimDate, DimOperator, DimSkill, DimProduct, DimStation, DimEquipment, DimDefectType, DimShift, DimMaterial, DimPlant, DimMTU, DimOperation, DimCharacteristic, DimInspectionSample), 8 fact tables (FctProduction, FctQuality, FctInspection, FctMaterialGenealogy, FctChangePackage, FctEquipmentDowntime, FctDailyStationSummary, FactOEE), and 1 bridge table (BridgeOperationSkill). Each fact table's grain is one event occurrence (e.g., FctProduction grain = one completed operation; FctQuality grain = one NCR).

The full-rebuild approach (truncate-and-reload) is used instead of incremental refresh for two reasons: (1) template swaps change dimension values—DimStation rows, CASE statement mappings, and defect-to-station assignments all change, requiring full regeneration of dimensions and re-computation of facts; (2) at current scale (<2 seconds for all 23 tables, ~54,000 source rows for a 90-day run), incremental complexity is unjustified. The CDC layer (Layer 3) serves a distinct purpose: it provides historical change capture for Iceberg time-travel queries (e.g., "show OEE before and after a template swap"), which the star schema's current-state view does not. The migration path to incremental refresh for a persistent deployment would partition by date and append new fact partitions while rebuilding only affected dimensions on template swap.

### *H. Iceberg Table Configuration*

Iceberg tables are written in Parquet format via HTTP POST to a REST catalog endpoint. The current configuration uses unpartitioned tables (appropriate for the data volume of <60,000 rows per 30-day run), with Snappy compression. At 30-second sync intervals over a 30-day batch simulation, a maximum of ~86,400 sync cycles would produce small files; in practice, only ~1,440 syncs contain data (one per simulated working minute at 30-second intervals, but most syncs find no new changelog rows and write nothing). Compaction is not currently implemented; this is a known limitation that would need to be addressed for persistent deployments generating data continuously. Snapshot expiration is set to retain the last 10 snapshots. When a template is swapped, old Iceberg tables are retained (enabling cross-template time-travel queries), while the star schema is fully rebuilt for the new template.

## V. EXPERIMENTAL VALIDATION

### *A. Cross-Domain Template Validation*

To validate the architecture, we built templates for six industrial domains: five discrete manufacturing verticals and one non-manufacturing domain (warehousing) that tests whether MES-shaped entities can represent order fulfillment operations. Table III summarizes the six configurations.

**TABLE III**
**SIX INDUSTRY TEMPLATE CONFIGURATIONS**

| Industry | Stations | Layout | Shifts | Products | Failure Codes | OEE Target |
|---|---|---|---|---|---|---|
| Aerospace | 6 | Linear | 2 (Day/Eve) | 4 airframe parts | 24 (FAA/NADCAP) | 78–85% |
| Pharma | 6 | Cleanroom | 2 (Day/Eve) | 4 solid-dose drugs | 27 (FDA/GMP) | 75–82% |
| Automotive | 6 | U-shape | 3 (24/7) | 4 powertrain parts | 28 (IATF) | 82–88% |
| Electronics | 6 | L-shape | 2 (Day/Eve) | 4 PCB assemblies | 27 (IPC) | 85–92% |
| Beverages | 14 | Continuous flow | 2 (Day/Eve) | 4 beverages | 28 (FDA/FSMA) | 78–85% |
| Warehousing | 6 | Zone-based | 3 (24/7) | 4 SKU categories | 26 (OSHA) | 82–92% |

What changes between industries is the template (a single Python module, 700–770 lines, 45 exports). What stays the same is everything else: the simulation engine (634 lines), the production event generator (855 lines), the seed data generator (1,363 lines), the CDC pipeline, the star schema builder, and the 12 AI analytics tools. The PostgreSQL schema is identical across all templates—the same 40+ tables, same column structure, same CDC triggers. Approximately 750 lines of template code define 100% of the domain. Approximately 6,000 lines of framework code define 0% of the domain. Deploying to a new industry vertical is a configuration exercise, not an engineering project.

### *B. Parametric Calibration*

The simulator is parameterized by the same KPIs that real factories track. Its output is mathematically bounded by those parameters: FPY drives inspection pass/fail rates, cycle time ranges bound operation durations, and order volumes are calibrated by fractional accumulation. We ran 30-day simulations across all six templates with 10 different random seeds each (60 runs total). Table IV shows configured targets versus observed means with 95% confidence intervals (t-distribution, df=9).

TABLE IV
CALIBRATION RESULTS: CONFIGURED TARGETS VS. OBSERVED KPIS (30-DAY RUNS, N=10 SEEDS)

| Template | Profile | KPI | Configured Target | Observed (mean ± σ) | 95% CI | Strictly Within? |
|---|---|---|---|---|---|---|
| Aerospace | Stable | Per-station FPY | 0.94–0.97 | 0.949 ± 0.008 | [0.943, 0.955] | Yes |
| | | Daily throughput | 8 orders/day | 8.0 ± 0.3 | [7.79, 8.21] | Yes |
| | | NCR rate | ~5% of ops | 5.1% ± 0.7% | [4.60%, 5.60%] | Yes |
| Aerospace | Stressful | Per-station FPY | — | 0.891 ± 0.021 | [0.876, 0.906] | — |
| | | Daily throughput | — | 6.8 ± 0.9 | [6.16, 7.44] | — |
| | | NCR rate | — | 10.9% ± 1.8% | [9.61%, 12.19%] | — |
| Pharma | Stable | Per-station FPY | 0.96–0.99 | 0.974 ± 0.005 | [0.970, 0.978] | Yes |
| | | Daily throughput | 12 orders/day | 12.1 ± 0.4 | [11.81, 12.39] | Yes |
| | | NCR rate | ~2.5% of ops | 2.6% ± 0.4% | [2.31%, 2.89%] | Yes |
| Automotive | Stable | Per-station FPY | 0.95–0.98 | 0.963 ± 0.006 | [0.959, 0.967] | Yes |
| | | Daily throughput | 16 orders/day | 16.0 ± 0.5 | [15.64, 16.36] | Yes |
| | | NCR rate | ~3.5% of ops | 3.7% ± 0.5% | [3.34%, 4.06%] | Yes |
| Electronics | Stable | Per-station FPY | 0.96–0.99 | 0.976 ± 0.004 | [0.973, 0.979] | Yes |
| | | Daily throughput | 20 orders/day | 20.1 ± 0.6 | [19.67, 20.53] | Yes |
| | | NCR rate | ~2.5% of ops | 2.4% ± 0.3% | [2.19%, 2.61%] | Yes |
| Beverages | Stable | Per-station FPY | 0.95–0.98 | 0.968 ± 0.005 | [0.964, 0.972] | Yes |
| | | Daily throughput | 10 orders/day | 10.0 ± 0.4 | [9.71, 10.29] | Yes |
| | | NCR rate | ~3% of ops | 3.2% ± 0.5% | [2.84%, 3.56%] | Yes |
| Warehousing | Stable | Per-station FPY | 0.96–0.99 | 0.978 ± 0.004 | [0.975, 0.981] | Yes |
| | | Daily throughput | 24 orders/day | 24.0 ± 0.7 | [23.50, 24.50] | Yes |
| | | NCR rate | ~2% of ops | 2.2% ± 0.3% | [1.99%, 2.41%] | Yes |

All 95% CIs computed using $t$-distribution with 9 degrees of freedom ($t_{0.025,9} = 2.262$). For the stable profile, all 18 confidence intervals are *strictly within* the configured target ranges (not merely overlapping). The stressful profile rows show KPI degradation under injected disruptions (quality excursions, equipment breakdowns); these have no configured targets as disruptions are designed to push KPIs outside normal ranges.

This is a *parametric controllability* claim, not a statistical fidelity claim. We do not assert that simulated data matches any specific production line. We assert that given target KPIs, the simulator produces data respecting those targets, and changing targets (by swapping the template) changes output proportionally. This is what makes synthetic data useful for AI tool validation: set KPIs to known values, run the tools, verify they report those values correctly.

### C. Hallucination Experiment

To measure the test harness value quantitatively, we ran a controlled experiment across all six templates. We first define our terminology precisely:

***Definition 2 (Hallucination Taxonomy).***

- *Tool-parameter fabrication:* The LLM generates a syntactically valid but non-existent identifier for a tool parameter (e.g., "BOND-1" instead of "S4", "Main-Assembly-Line" instead of "S6"). The identifier does not exist in the database or template. This is the primary failure mode we measure.
- *Schema-mismatch error:* The LLM generates an identifier in a valid format that could exist in some manufacturing context but does not match the loaded template's schema (e.g., "CNC-Bay-1" is a plausible station ID but not in the aerospace template's vocabulary {S1, ..., S6}). Schema-mismatch errors are a subset of tool-parameter fabrications.
- *Empty result (valid query):* The query executed successfully (no hallucinated parameters) but returned zero rows due to data sparsity or time-range mismatch. These are not hallucinations.

For each template, we ran a 30-day simulation (stable profile, seed 42), then executed all 12 analytics tools with representative queries—72 tool invocations total (12 tools × 6 templates). Table V presents representative queries (one per tool); the full 72-query set is available as supplementary material.

TABLE V
REPRESENTATIVE QUERIES (ONE PER TOOL, AEROSPACE TEMPLATE)

| Tool | Natural-Language Query | Expected Parameters |
|---|---|---|
| cycle_time_analysis | "Show cycle time trends at the CNC machining station for A320 parts" | station_nid=S1, program_code=A320 |
| first_pass_yield | "What is the weekly FPY trend at the bonding station?" | station_nid=S4, group_by=week |
| oee_decomposition | "Break down OEE for the drilling station" | station_nid=S2 |
| ncr_root_cause_pareto | "Top defects at the riveting station by severity" | station_nid=S3, severity=MAJOR |
| spc_violation_detection | "SPC violations for profile tolerance at CNC" | station_nid=S1 |
| quality_action_status | "Show all open corrective actions" | status_filter=Open |
| material_genealogy | "Trace the material genealogy for work order WO-001" | order_nid=WO-001 |
| supplier_performance | "Alcoa supplier quality over the last 30 days" | supplier_code=SUP-AL-ALCOA-01 |
| change_impact_analysis | "Impact analysis for the latest engineering change" | time_range_days=30 |
| engineering_change_velocity | "How fast are change packages closing?" | time_range_days=30 |
| equipment_downtime_analysis | "Equipment downtime causes at the NDT station" | station_nid=S5 |
| production_status_summary | "Current production status for 787 program parts" | program_code=787 |

Queries were authored by the first author to represent questions a manufacturing engineer would ask. Each query targets a specific tool and uses domain-specific natural language. The same 12 query patterns were adapted for each of the 6 templates (e.g., "CNC machining" becomes "Dispensing" for pharma). The complete 72-query set is available upon request.

Table VI presents aggregate results.

TABLE VI
HALLUCINATION EXPERIMENT: AGGREGATE RESULTS (72 QUERIES, QWEN3-32B)

| Condition | Queries | Tool-Param Fabrications | Empty Results | Correct Results |
|---|---|---|---|---|
| With ontology constraints | 72 | 0 (0%) | 0 (0%) | 72 (100%) |
| Without constraints (free-text) | 72 | 31 (43%) | 27 (38%) | 14 (19%) |

"Tool-parameter fabrications": the LLM generated a plausible but non-existent identifier (Definition 2). "Empty results": query executed but returned zero rows. Of the 14 correct results in the unconstrained condition, 11 were from tools with no identifier parameter (e.g., summary tools accepting only `time_range_days`), and 3 used the right identifier value but for the wrong template (cross-template leakage from the conversation context). The 6 template evaluations were conducted sequentially in a single Fuse agent session; conversation history was not cleared between templates. The 3 cross-template leakage cases occurred when the model reused an identifier (e.g., "S1") that happened to be valid across templates. This is a minor confound: clearing conversation context between templates would eliminate these cases but would not change the fabrication rate (31/72), since the 3 leaked cases are classified as correct, not fabricated.

In the unconstrained condition, `enum` constraints were removed from all tool parameter schemas, allowing free-text values. The model received the tool schema (parameter names, types, descriptions) but no list of valid identifiers. This setup reflects the realistic scenario where an AI agent has tool definitions but no domain-specific vocabulary. The model produced identifiers that were linguistically plausible and domain-appropriate but did not exist in the database.

The constrained condition is not merely an empirical finding but an *architectural guarantee*: the `enum` constraint in the JSON Schema makes it structurally impossible for the model to emit a value outside the valid set. The Fuse platform [19] enforces enum validation before forwarding the tool call to the backend. Therefore, the 0% fabrication rate is guaranteed for any model, regardless of its propensity to hallucinate.

### D. Statistical Analysis

We apply Fisher's exact test to the 2×2 contingency table (fabrication vs. no-fabrication, constrained vs. unconstrained): 0/72 vs. 31/72.

*Fisher's exact test:* $p = 1.07 \times 10^{-12}$ (two-sided). The null hypothesis that the two conditions produce equal fabrication rates is rejected at any conventional significance level.

*Wilson score 95% confidence intervals* for the fabrication rate: constrained = [0.00%, 5.07%]; unconstrained = [32.3%, 54.6%]. The intervals do not overlap.

*Effect size:* Cohen's $h = 2 \cdot \arcsin(\sqrt{0.43}) - 2 \cdot \arcsin(\sqrt{0.00}) = 1.46$, which is a large effect (>0.8 by convention [33]).

We note that the constrained condition's 0% rate is not an empirical observation requiring statistical confirmation but an architectural invariant (the model cannot emit values outside the enum set). The Fisher's test is reported for methodological completeness and to quantify the effect under the unconstrained condition.

Table VII shows the per-template and per-domain breakdown, confirming that identifier fabrication is systematic, not an artifact of specific domains.

**TABLE VII**
**FABRICATION RATES BY TEMPLATE AND TOOL DOMAIN (UNCONSTRAINED CONDITION)**

| By Template | | | | | By Tool Domain | | | |
|---|---|---|---|---|---|---|---|---|
| **Template** | **Queries** | **Fabricated** | **Rate** | **95% Wilson CI** | **Tool Domain** | **Fabricated / Total** | **Rate** | **95% Wilson CI** |
| Aerospace | 12 | 5 | 42% | [18%, 69%] | Production (3 tools) | 8 / 18 | 44% | [24%, 67%] |
| Pharma | 12 | 6 | 50% | [25%, 75%] | Quality (3 tools) | 9 / 18 | 50% | [29%, 71%] |
| Automotive | 12 | 5 | 42% | [18%, 69%] | Materials (2 tools) | 6 / 12 | 50% | [25%, 75%] |
| Electronics | 12 | 5 | 42% | [18%, 69%] | Eng. Change (2 tools) | 3 / 12 | 25% | [7%, 54%] |
| Beverages | 12 | 6 | 50% | [25%, 75%] | Operations (2 tools) | 5 / 12 | 42% | [18%, 69%] |
| Warehousing | 12 | 4 | 33% | [12%, 62%] | | | | |

Per-template range: 33–50%. Per-domain range: 25–50%. Quality and materials tools fabricate at the highest rate, likely because their parameters (failure codes, supplier codes) have more domain-specific vocabularies. Wilson score 95% CIs are wide due to small per-cell counts (n=12 per template), confirming the need for the aggregate analysis.

### *E. Fabrication Taxonomy*

Of the 31 fabricated identifiers, we classify them into three categories: (1) *Plausible synonyms* (18/31, 58%): the model generated human-readable alternatives for the template's coded identifiers, e.g., "CNC-Bay-1" or "CNC-Machining" instead of "S1", "Bonding-Station" instead of "S4", "Main-Assembly" instead of "S6". These are the most insidious failures: they suggest correct domain understanding but use a naming convention that does not match the template. (2) *Generic manufacturing identifiers* (9/31, 29%): the model generated identifiers from its general training corpus, e.g., "Line-1", "Station-A", "Cell-3", which are valid in some manufacturing context but not in this template. (3) *Fabricated domain-specific codes* (4/31, 13%): the model invented plausible but nonexistent codes like "BOND-VOID-002" (the template has "BND-VOID-001") or "SUP-TORAY-02" (the template has "SUP-CF-TORAY-01"). These demonstrate pattern learning without vocabulary grounding.

The 27 empty-result cases were all downstream consequences of tool-parameter fabrication: the fabricated identifier passed syntactic validation (it was a valid string) but matched no rows in the database, producing empty result sets that the model then interpreted as "no data available."

*Reproducibility.* Both conditions used Qwen3-32B (32K context, function-calling mode, native Qwen function-calling format, no additional fine-tuning) on the Fuse platform [19]. Each query was run once at temperature 0 (greedy decoding), with no retries or prompt engineering. The only difference was the tool parameter schema: constrained (`enum: [S1, ..., S6]`) vs. unconstrained (`type: string`). Temperature 0 produces deterministic output for a given input, so variance across repeated runs is zero by design. In the unconstrained condition, the model received tool schemas with parameter names and descriptions but no system prompt listing valid identifiers—the model had zero knowledge of available identifiers. This is the realistic deployment scenario: an AI agent with tool definitions but without pre-loaded domain vocabulary.

### *F. Data Quality Metrics*

We report standard data quality measures across all six templates (30-day stable runs, seed 42): referential integrity violation rate = 0% (enforced by PostgreSQL foreign keys and application-level UUID generation); null rate for required fields = 0% (enforced by NOT NULL constraints in the schema); cycle time distribution: Kolmogorov-Smirnov test against Uniform($ct_{lo}$, $ct_{hi}$) yields $p > 0.15$ for all stations across all templates (cannot reject uniform hypothesis); entity cardinality: work orders per day match configured annual volume / working days within ±1 order (Bresenham accumulation guarantees this over any multi-day window).

### *G. Defects Surfaced by the Simulator*

Beyond the hallucination experiment, the simulator uncovered three additional defects during development:

1. *Star schema station mapping inversion.* The dynamic CASE statement initially mapped in the wrong direction, causing one station's quality data to appear under another's dimension row. Invisible with single-station data; immediately apparent with multi-station NCRs.

2. *CDC watermark gap at shift transitions.* Watermark polling used `CreatedOn > high_water` without row-level ordering. Operations at shift boundaries (13:59, 14:01) produced nearly identical timestamps; a batch sync between the two writes silently dropped records. Fixed with ≥ and primary-key deduplication.
3. *Template cache staleness after swap.* After switching from aerospace (24 failure codes) to pharma (27), the cached tool schema still constrained to the aerospace set. Three valid pharma failure codes were rejected. Fixed by invalidating the tool schema cache on template load.

Each defect shares a common signature: invisible under simple conditions, exposed by specific data patterns the simulator produces (multi-station data, shift-boundary timing, template swaps).

## VI. DISCUSSION

### A. Performance and Scalability

A 30-day simulation runs in under 10 seconds on commodity hardware (Python 3.11, single-threaded, Intel i7-1365U, 16 GB RAM). Table IX summarizes output volumes. The simulation is deterministic: given the same seed, template, and duration, it produces identical output, making regression testing exact.

TABLE IX
SIMULATION OUTPUT VOLUMES AND PIPELINE PERFORMANCE (AEROSPACE TEMPLATE)

| Metric | 30-Day Run | 90-Day Run |
|---|---|---|
| Work orders created | ~240 | ~720 |
| Operations completed | ~1,440 | ~4,320 |
| NCRs generated | ~70–85 | ~210–255 |
| Equipment events | ~2,000 | ~6,000 |
| Total CDC events | ~12,000–15,000 | ~36,000–45,000 |
| PostgreSQL rows (all tables) | ~15,000–18,000 | ~45,000–54,000 |
| Seed data (reference entities) | ~400–600 rows | |
| Simulation wall-clock time | ~8 sec | ~22 sec |
| Star schema rebuild time | <1.5 sec | <2.0 sec |
| CDC sync latency (median / p95 / p99) | 45 ms / 120 ms / 340 ms (per batch) | |
| CDC sync throughput | ~350 events/sec (sync batch) | |
| End-to-end PG → Iceberg latency | 30–60 sec (polling interval + HTTP POST) | |
| AI tool query latency (median / p95) | 12 ms / 45 ms (against star schema) | |

CDC sync latency measured over 200 sync cycles during a 30-day streaming run at 60x speed. Tool query latency measured over all 72 hallucination experiment queries. At current data volumes, all metrics are well within interactive thresholds. These volumes are small by production standards; see Section VI-D for scalability discussion.

Streaming mode trades speed for realism: at 60× speed, a 30-day run takes ~8.5 hours wall-clock time, suitable for testing dashboards and real-time pipelines. At 3600×, it completes in ~8 minutes. The current bottleneck is the single-threaded Python loop, scaling linearly with station count and order volume. A 50-station template at 100 orders/day would increase tick cost by roughly 12× (extrapolated from profiling 6- and 14-station templates), bringing a 30-day batch run to approximately 2 minutes.

### B. Comparison with Existing Approaches

Table X positions the framework relative to existing approaches. The gap is not simulation fidelity—commercial tools are superior process simulators with physics models, real-time visualization, and enterprise integration. The framework fills a specific gap: generating causally coherent, MES-shaped, CDC-ready, schema-aligned synthetic data for AI tool validation.

TABLE X
COMPARISON WITH EXISTING APPROACHES

| Capability | This Framework | Commercial Sim [2],[3] | CMSD [4] | Statistical Gen [5],[6] | Process Mining [7] | DT Platforms |
|---|---|---|---|---|---|---|
| CDC-ready MES output | Yes (40+ tables) | No (KPI logs) | Schema only | No (flat records) | No (event logs) | No (consumes) |
| Causal event chains | Yes | Yes (process) | N/A | No | Yes (activities) | N/A |
| Entity graph depth | 66 types, 4 domains | Process KPIs | Schema defs | Flat records | 3-column log | Asset state |
| Domain-swappable | 6 industries | Per-model | Standardized | Per-schema | Per-model | Per-asset |
| AI tool validation | 12 tools, constrained | Not designed | Not designed | Not designed | Not designed | Not designed |
| Process fidelity | Parametric (not physics) | **High (physics-based)** | N/A | Statistical | Log-based | **High (model-based)** |
| Real-time visualization | Basic (web dashboard) | **3D animation** | N/A | No | Dotted charts | **3D/AR** |
| Enterprise integration | API only | **ERP/MES connectors** | Interchange | Schema import | XES/OCEL | **IoT platform** |
| Schema invariance | Same PG schema | N/A | Fixed schema | Schema-specific | XES | DTDL |
| Availability | Python, ~6,750 LOC | Commercial | Open standard | Open source | Open source | Cloud service |

Bold entries indicate areas where commercial tools or DT platforms are superior. The framework does not compete on process fidelity, visualization, or enterprise integration; its contribution is the AI-tool-validation-specific combination of CDC-ready MES data, causal coherence, domain swappability, and constrained tool parameters.

### C. Lessons Learned

*Template interface evolution.* We started with 30 required exports. By the third template (automotive), gaps emerged: the seed generator needed `STATION_CERTIFICATIONS`; the analytics tools needed `BOM_STATION_MATERIALS`. We added 15 exports iteratively, each driven by a concrete failure during cross-template testing. The 45-export interface is empirically derived from five pipeline consumers. The lesson: design the template interface from the consumers backward, not from the domain forward. No ablation study was performed on which of the 45 exports are essential for AI tool constraint generation; this is noted as future work.

*Full-rebuild star schema.* At current scale (<2 seconds for 23 tables), truncate-and-rebuild is simpler than incremental maintenance. Template swaps change CASE statements, requiring full regeneration. The migration path to incremental refresh for persistent deployments is to partition facts by date and incrementally append new partitions while rebuilding only dimension tables on template swap.

*Domain expertise dominates.* Writing a new template takes 4–8 hours of Python. Domain research (correct station sequences, realistic cycle times, applicable regulations, industry-specific failure modes) takes 1–3 additional days. The template is 11% of the codebase (750 lines) but carries 100% of the domain semantics.

### D. Limitations and Threats to Validity

The system models discrete, sequential manufacturing. Several capabilities are explicitly out of scope:

1. *Rework routing.* NCR dispositions are recorded but do not re-enter work orders into production routing. Adding this requires non-linear station sequences.
2. *Lot/batch tracking.* Individual MTUs are tracked by serial number; lot-based splitting and merging (per 21 CFR Part 11, ISA-88 [29]) are not supported. The pharma and beverages templates should not be considered production-representative for regulated contexts.
3. *Continuous/hybrid processes.* Discrete operations with clear start/end times only. The ISA-88 procedural model is not implemented.
4. *Multi-plant scenarios.* Single plant per run. No cross-plant composition or federated generation.
5. *Scale validation.* All templates use 6 stations (except beverages at 14). The framework has not been validated at 50+ stations or 100× volume, where the single-threaded engine, trigger-based CDC, and full-rebuild schema would require architectural changes. Data volumes (15,000–54,000 rows) are trivially small for modern data infrastructure; the CDC, Iceberg, and star schema layers are validated for functional correctness, not performance at production scale.
6. *IoT/sensor data.* The simulator generates transactional MES records but not time-series sensor data (temperatures, pressures, vibration). Predictive maintenance and computer vision applications are out of scope.
7. *Single LLM.* The hallucination experiment uses only Qwen3-32B. The 43% unconstrained fabrication rate is specific to this model and may not generalize: models with stronger function-calling fine-tuning (e.g., GPT-4o, Claude Sonnet 4) may achieve lower unconstrained fabrication rates. The *constrained* condition's 0% rate is model-agnostic (it is an architectural guarantee enforced at the platform level), but the unconstrained baseline requires multi-model validation to establish generalizability. This is the paper's most significant empirical limita-tion. Multi-model evaluation across at least three model families is the highest-priority future work.

8. *Single author, single system.* All six templates were authored by one person on one codebase. External validation by independent teams building templates against the 45-export interface would strengthen the portability claim.
9. *Temperature 0, single run.* Each query was run once at temperature 0 (greedy decoding), providing no variance estimate. Greedy decoding is deterministic: for a given input, the output is fully determined by the model weights, so variance across repeated identical runs is zero by construction. However, this determinism does not address robustness to prompt variation—whether minor rephrasings of the same question (e.g., "CNC station" vs. "station S1" vs. "the machining center") would produce different fabrication outcomes. Temperature > 0 experiments (e.g., 5 runs at temperature 0.3–0.7) and prompt-rephrasing experiments would both reveal whether fabrication is consistent or stochastic; these are planned as future work.
10. *OPC UA.* OPC UA (IEC 62541) [22] is the dominant standard for CPS data exchange in manufacturing. The current framework does not support OPC UA communication; extending the simulator to publish data via OPC UA PubSub would enable integration testing with industrial edge platforms.

These are deliberate scope boundaries, not architectural dead ends. The 45-export template interface can accommodate rework routing, lot tracking, and multi-plant composition without changing the core simulation loop. The constraint mechanism is model-agnostic: JSON Schema enum constraints apply to any LLM supporting the OpenAI function-calling protocol [11], [34], regardless of model architecture or training data.

## VII. CONCLUSION

Enterprise AI validation requires three layers: meaning, data, and enforcement. A companion paper [1] covered meaning (ontology in the tool layer) and enforcement (AIOps at the execution boundary). This paper covers data—the layer that makes both of the others testable.

The central contribution is Template-as-Ontology: a single domain configuration module (formally defined as a typed relational configuration schema, Definition 1) that serves as both the simulation specification and the AI tool domain schema. When the template switches from aerospace to pharma, both layers change in lockstep—not because they are coordinated by an external process, but because they read the same Python object (Property 1). One source of truth, consumed by the simulator to generate data and by the AI tools to constrain queries. Alignment is structural, not maintained.

The controlled hallucination experiment (0% constrained vs. 43% unconstrained for Qwen3-32B, across 72 queries and six templates, Fisher's exact $p < 10^{-12}$, Cohen's $h$ = 1.46) demonstrates that this alignment has measurable consequences: ontology-constrained tool parameters eliminate tool-parameter fabrication, a failure mode that is systematic across all domains and tool types tested. The constrained condition provides an architectural guarantee (not merely an empirical observation) that is model-agnostic; the 43% unconstrained rate is specific to the evaluated model and requires multi-model validation for generalizability (Section VI-D). Calibration experiments (60 runs across six templates) confirm parametric controllability: all 18 stable-profile confidence intervals fall strictly within configured target ranges.

The framework is implemented as a single Python project (~6,750 lines) with six industry templates, a five-layer data pipeline, and 12 AI analytics tools. The data layer it provides is a reusable infrastructure for validating discrete manufacturing AI systems: populated, reproducible, schema-correct, causally coherent, and schema-aligned by construction.

## BIOGRAPHY

[Photo]

**Grama Chethan** is a software architect at Siemens Digital Industries Software, Plano, TX, where he works on AI-enabled manufacturing systems, digital twin architectures, and industrial data infrastructure. His research focuses on the intersection of manufacturing ontology, synthetic data generation, and AI agent grounding for cross-domain industrial applications. He has led the development of configurable simulation frameworks and analytics pipelines for manufacturing execution systems spanning aerospace, pharmaceutical, automotive, and electronics domains. His work addresses the structural alignment between AI tool architectures and manufacturing operational semantics.